\documentclass[runningheads]{llncs}

\usepackage{graphicx}
\usepackage[lofdepth,lotdepth]{subfig} 
\usepackage{multirow}

\usepackage{booktabs}					

\usepackage{hyperref} 				

\usepackage{amsmath}
\usepackage{amsfonts}
\usepackage{amssymb}
\usepackage{commath}

\usepackage[]{algorithm2e}
\SetKwInOut{Data}{Data}
\SetKwInOut{Model}{Model}
\SetKwInOut{Parameters}{Parameters}

\begin{document}

\title{ptype-cat: Inferring the Type and Values of Categorical Variables}

\author{Taha Ceritli\inst{1,2} \and\\
Christopher K.~I.~Williams\inst{1,2}}

\institute{
School of Informatics, University of Edinburgh, UK 
\and 
Alan Turing Institute, London, UK
}
\maketitle         
\begin{abstract}
Type inference is the task of identifying the type of values in a data column and has been studied extensively in the literature. Most existing type inference methods support data types such as Boolean, date, float, integer and string. However, these methods do not consider non-Boolean categorical variables, where there are more than two possible values encoded by
integers or strings. Therefore, such columns are annotated either as integer or string
rather than categorical, and need to be transformed into categorical
manually by the user. In this paper, we propose a probabilistic type inference method that can identify the general categorical data type (including non-Boolean variables). Additionally, we identify the possible values of each categorical variable by adapting the existing type inference method \emph{ptype}. Combining these methods, we present \emph{ptype-cat} which achieves better results than existing applicable solutions.

\keywords{Data dictionary \and Type inference \and Categorical variables}
\end{abstract}

\section{Introduction}
\makeatletter{\renewcommand*{\@makefnmark}{}
\footnotetext{Presented at the ECML-PKDD Workshop on Automating Data Science, 17 Sept 2021.}\makeatother}

A \emph{data dictionary} is defined as ``a centralized repository of information about data such as meaning, relationships to other data, origin, usage, and format'' \cite{mcdaniel1994ibm}. By supporting the analyst in gaining insights into the data, the data dictionary plays a central role in the entire process of data analytics. A core component commonly reported in data dictionaries is the \emph{data type} which specifies the type of values in a data column. The data type (such as integer or string) is one of the fundamental properties of a data column that needs to be understood before further analysis. In this work, we consider five main data types: categorical, date, float, integer
and string (see Appendix \ref{sec:defs-types} for a detailed description of these data types) and tackle the task of associating a column of data with one of these types.

Automatic identification of the categorical type from the data in a column is particularly challenging because the values may be \emph{encoded} as \emph{strings} or \emph{integers}. For example, in a data table about clothing, a variable ``Class Name" could be a categorical variable taking on values such as \texttt{Jackets}, \texttt{Dresses} and \texttt{Pants},  while a variable ``Rating'' may take on values in a fixed range \texttt{1} through \texttt{5}.\footnote{
We include ordinal variables (as in this example) as
categorical variables, although it may be useful to distinguish them further in subsequent analysis.} To the best of our knowledge, these issues are not addressed by any existing work in the literature, except Bot (proposed by Majoor and Vanschoren \cite{majoor2018}), OpenML and Weka which tackle the type inference based on heuristics such as labeling a column as categorical when the number of unique values is lower than a threshold (see Sec.\ \ref{sec:related_work} for a detailed discussion). In this work, we use machine learning rather than heuristics to infer the type of a data column, and show that our probabilistic approach can be more flexible than hard-choices made with heuristics. Our contributions can be summarized as follows: 

\begin{itemize}
\item We propose a predictor that can identify the data type (categorical, date, float, integer and string) for each column of a dataset (Section \ref{sec:methodology}).
\item We define \emph{inference of categorical values} as \emph{the task of identifying the possible values a categorical variable can take on}. We address this task by adapting \emph{ptype} \cite{ceritli2020ptype} which can robustly determine the possible values of a categorical variable by identifying missing data and anomalies in a data column (Section \ref{sec:methodology}).
\item We show that the our methods outperform the existing methods using a large number of datasets (Section \ref{sec:experiments}).
\end{itemize}

\section{Methodology}
\label{sec:methodology}

\paragraph{\bf{Background:}} In this work, we extend the probabilistic type inference method called ptype \cite{ceritli2020ptype}. Assuming that the data entries are read as strings, ptype allows us to infer a plausible column type (Boolean, date, float, integer or string) for a data column, and, conditioned on that type, identify any values which are deemed missing or anomalous. ptype uses Probabilistic Finite-State Machines (PFSMs) to model data types including known column types, missing and anomaly types. Combining these PFSMs, it provides a type inference method that outperforms the existing methods by its ability to detect missing and anomalous entries in a data column. Given a data column, ptype can be used to calculate the posterior distribution of the column type, where the maximum posterior probability denotes the most likely data type for that column. We could na\"{i}vely apply ptype to our task by mapping the Boolean type to the categorical type, but this would not correctly handle non-Boolean categorical variables. 

\paragraph{\bf{Our Proposed Model:}} Our goal here is to obtain the posterior probability distribution of column type over the categorical, date, float, integer and string types, which is achieved in two steps. 
Initially, assuming that a column of data $\textbf{x}=\{x_i\}_{i=1}^N$ has been read in where each $x_i$ denotes the characters in the $i^{th}$ row and $N$ is the number of rows in a data column, we calculate the posterior probability distribution $p(t|\bf{x})$ of column type $t$ over the date, float, integer and string types by running a modified form of ptype that excludes the Boolean type. If a data column is labelled with the date or float type according to this posterior probability distribution, we assume that the posterior probability for the categorical type is zero and use the distribution as it is. Otherwise, if a data column is labelled with the integer or string type, we employ a separate binary classifier to determine the posterior probability for the categorical type, where the initial posterior probabilities for the four types are treated as features. The resulting method is called ptype-cat.

Note that we discard the Boolean feature used in the default setting of ptype as it leads to a limited capability to detect the categorical type. Instead, we propose four new features to characterize the categorical type. Two of our proposed features are the number of unique values in a data column and the uniqueness ratio, which are respectively denoted by $U$ and $R$ where $R$ is defined as $U/N$. We extract the same features by taking into account the ``clean'' entries of a data column rather than all the data entries. Note that the clean entries refer to the data entries which are neither missing nor anomalous. These features are respectively denoted by $U_{c}$ and $R_{c}$, where $R_{c}$ is defined as $U_{c}/N_{c}$. Therefore, we obtain 8 features after combining ptype features with ours. 

When a data column is labelled with the integer or string type, we determine the posterior probability for the categorical type by re-distributing the probability mass for the integer or string type according to a trained model, e.g., if the data type is initially inferred as integer, we divide the posterior probability for the integer type between the integer and categorical types according to a trained model as shown in Fig.\ \ref{fig:ptype-cat-split} (Supp.\ Mat.). To train the binary classifier for categorical/not-categorical classification, we use data columns annotated as integer, string and categorical. Mapping the integer and string labels to not-categorical, we train a binary Logistic Regression via 5-fold nested cross-validation, where we estimate its hyperparameters through grid-search. Note that the hyperparameter of Logistic Regression and the corresponding range of values used in the grid-search are reported in Appendix \ref{sec:add-exp-results}.

\paragraph{\bf{Identification of Categorical Values:}} The task here is to infer the possible values a given categorical data column can take on. A na\"{i}ve approach would be to treat all the unique values in a column as the corresponding categorical values. However, this method, which we call Unique, would fail when the data contains missing and anomalous values, as such values would be labeled as valid categorical values rather than being discarded. To address this problem, we employ ptype which labels the entries of a data column that are not missing and anomalous as ``clean''. We treat these clean entries as the categorical values of a data column. ptype can be used to calculate the corresponding posterior probability of row type being ``clean'' for each unique entry.

\section{Related Work}
\label{sec:related_work}
Existing type inference methods, including F\# \cite{fsharp-data-pldi2016Short}, hypoparsr \cite{hypoparsr2017}, messytables \cite{messytables2017}, PADS \cite{fisher2005pads}, ptype \cite{ceritli2020ptype},  readr \cite{readr2017}, Test-Driven Data Analysis (TDDA) \cite{tdda2018} and Trifacta \cite{Trifacta2018}, do not support non-Boolean categorical variables and therefore would have limited performance on our task. 

The closest related work to ours is Bot proposed by Majoor and Vanschoren \cite{majoor2018}. Bot reads the data by using the Pandas.read\_csv() function and applies a set of heuristics to map the inferred data types to the types considered in this work. For example, a data column initially labelled as integer is treated as a categorical variable when one of the following conditions is satisfied: (i) if the number of unique values is less than 11 and (ii) if the number of unique values is between 10 and a pre-defined value and the average of absolute distances between integers is lower than the average of integers. These heuristics would help to identify categorical variables in certain scenarios. However, a trained machine learning model can offer a more flexible solution by learning the relationship between data features and the types.

An alternative to Bot is the methods used to convert Comma-separated values (CSV) files to the Attribute Relation File Format (ARFF)\footnote{The details are available at https://waikato.github.io/weka-wiki/formats and processing/arff stable/ [Accessed on 05/12/2020]}. For example, csv2arff\footnote{The code is available at
\url{https://github.com/openml/ARFF-tools/blob/master/csv-to-arff.py}
[Accessed on 09/11/2020]} is used in OpenML to identify ARFF data types (date, nominal, numeric and string) and categorical values. It uses the Pandas library to parse a data file and employs a rule-based approach that is similar to Bot. Therefore, this method also offers limited capability compared to a trained probabilistic model. Weka\footnote{The details are at \url{https://waikato.github.io/weka-wiki/formats_and_processing/converting_csv_to_arff/}[Accessed on 05/12/2020].} provides another rule-based approach for CSV to ARFF conversion, which suffers from similar issues as OpenML's converter.

\section{Experiments}
\label{sec:experiments}
We first describe our experimental setup (Sec.~\ref{sec:exp-setup}) and then present quantitative results on type inference and identification of categorical values (Sec.~\ref{sec:exp-res}). 

\subsection{Experimental Setup}
\label{sec:exp-setup}
We briefly describe the datasets, evaluation metrics and methods used in our experiments below. See Appendix \ref{sec:additional-exp-setup} for a detailed description. 

\underline{Datasets:} For type inference, we have used 86 datasets obtained from various sources such as Kaggle\footnote{\url{https://www.kaggle.com/datasets}
[Accessed on 18/11/2020]}, OpenML\footnote{\url{https://www.openml.org/search?type=data}
[Accessed on 18/11/2020]} (randomly selected through the API) and
UCI\footnote{\url{https://archive.ics.uci.edu/ml/datasets.php}
[Accessed on 18/11/2020]}. We have annotated each dataset in terms of data types and categorical values by hand, based on the available meta-data and the unique values in each data column. The data files, their sources and our annotations can be accessed via \url{https://bit.ly/2Ra2Vu7}. As we mention in Sec.~\ref{sec:methodology}, we apply nested cross-validation for hyperparameter selection, which is summarized in Algorithm \ref{alg:nested-cv} (Supp.\ Mat.). Once the datasets are split using 5-fold nested cross-validation, we collect their columns in the corresponding folds. Note that we split at the dataset level in order to avoid bias in test data. Our datasets contain a total number of 2989 columns (900 categorical, 49 date, 1462 float, 513 integer and 65 string). The number of categorical values in a column is between 1 and 80.

\underline{Evaluation Metrics:} To assess type inference, we use the metrics which are used to evaluate ptype, namely overall accuracy and the Jaccard index. Additionally, we plot the Precision Recall (PR) curve for each method and report the Average Precision (AP) of each curve. To assess identification of categorical values, we first evaluate the methods using overall accuracy. In addition, we calculate the Jaccard index per data column $J(A,B)$ defined as $|A \cap B| / |A \cup B|$ where $A$ and $B$ respectively denote the sets of annotated and predicted categorical values, and report their average over columns. 

\underline{Methods:} We compare ptype-cat with Bot, OpenML and Weka. As OpenML and Weka use the numeric type in the ARFF format, they do not distinguish integers and floats. To compare with ptype-cat and Bot below, we enhance them with ptype's functionality to make this call. In addition, we construct a baseline called Unique for inferring categorical values. Unique treats all unique values in a column as categorical values. Our code is publicly available at \url{https://github.com/tahaceritli/ptype-cat-experiments} for reproducibility. We have also incorporated ptype-cat into the Python package of ptype which is available at \url{https://github.com/alan-turing-institute/ptype}.

\subsection{Results}
\label{sec:exp-res}

\underline{Type Inference:} Table \ref{table:evals} presents the performance of the methods in terms of overall accuracy and Jaccard index. These results indicate that ptype-cat consistently outperforms the competitor methods, except for the float type where it performs very similarly to Weka. These improvements are generally thanks to the flexibility of our probabilistic approach, i.e., we train a probabilistic model to learn the relationship between data features and types, whereas the others employ certain heuristics to identify types (see Sec.\ \ref{sec:related_work} for a detailed discussion). 

Fig.\ \ref{fig:precision_curves} presents the PR curves obtained by the methods, which indicates a similar trend as above in that ptype-cat performs better than the other methods. Note that the competitor methods provide only either 0 or 1 as a score for each data type. In contrast, ptype-cat generates more fine-grained scores valued between 0 and 1.

The leading competitor method on type inference is  OpenML. Comparied to OpenML, our method correctly classifies an additional 141 data columns (62 categorical, 27 date, 39 float, 8 integer and 5 string). The main difference is that ptype-cat correctly classifies 51 categorical columns which are misclassified as integer by OpenML. We explain these differences in Appendix \ref{sec:add-exp-results}. In addition, McNemar's test (see e.g.,  \cite{dietterich1998approximate}) applied to the column type predictions of ptype and OpenML confirms that the methods are statistically significantly different from each other (see Appendix \ref{sec:add-exp-results-type-inference} for the details).

\begin{minipage}{\textwidth}
  \scalebox{0.8}{
    \begin{minipage}[b]{.5\textwidth}

    \centering
    \captionsetup{type=table} 
    \begin{tabular}{l|cccc}
\toprule
\multicolumn{1}{c}{} &  \multicolumn{4}{c}{\bfseries Method} \\\cline{2-5} \addlinespace[1mm]
  &  Bot & OpenML & Weka & ptype-cat \\
\midrule
Overall  & \multirow{2}{*}{0.84} & \multirow{2}{*}{0.88} & \multirow{2}{*}{0.79} & \multirow{2}{*}{\textbf{0.93}} \\
Accuracy & & \\
\hline
\addlinespace[1mm]
Categorical &   0.68 &              0.81 &            0.43 &        \textbf{0.85} \\
Date            &   0.08 &              0.00 &            0.00 &        \textbf{0.51} \\
Float &   0.83 &              0.95 &            \textbf{0.97} &        \textbf{0.97} \\
Integer &  0.64 &              0.59 &            0.49 &        \textbf{0.70} \\
String          &  0.29 &              0.20 &            0.06 &        \textbf{0.52} \\
\bottomrule
\end{tabular}
      \caption{Performance of the methods using
      the overall accuracy and per-class Jaccard index, for the
      Categorical, Date, Float, Integer and String types.}
      \label{table:evals}      
    \end{minipage}
    }
    \hfill
  \begin{minipage}[b]{0.5\textwidth}
    \centering
\includegraphics[width=\textwidth]{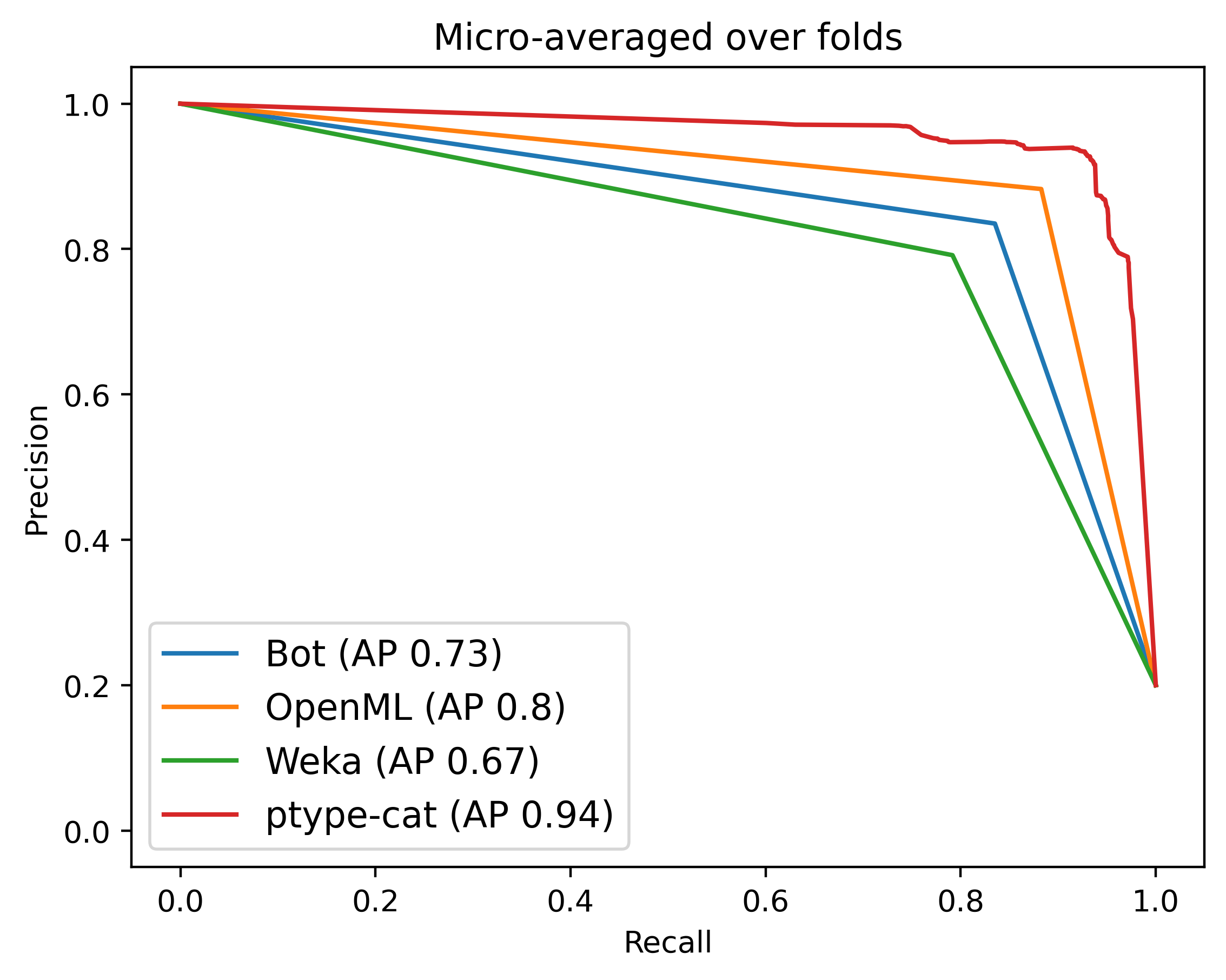}
    \captionof{figure}{PR curves for the methods.}
    \label{fig:precision_curves}
  \end{minipage}
  \end{minipage}
\vspace{1cm}

\underline{Identification of Categorical Values:} Table \ref{table:categorical_encoding} presents the performance of the methods on identification of categorical values. These results indicate that ptype-cat outperforms the competitor methods in terms of both metrics, and the leading competitor method is OpenML. We also observe that Unique performs better than Bot and WEKA, which produce similar results. 

\begin{table}[h!]
\centering
\begin{tabular}{l|ccccc}
\toprule
\multicolumn{1}{c}{} &  \multicolumn{5}{c}{\bfseries Method} \\\cline{2-6}
  & Bot  & OpenML & Weka & Unique & ptype-cat \\
\midrule
Overall  Accuracy &  0.33  & 0.83 &  0.38 & 0.64 & \textbf{0.90} \\
Average Jaccard & 0.40 &   0.87 &  0.42 & 0.87 & \textbf{0.92} \\
\bottomrule
\end{tabular}
\vspace{1 mm}  
\caption{Performance of the models on inference of categorical values.}
\label{table:categorical_encoding}
\end{table}
\vspace{-1cm}
The accuracies indicate that OpenML identifies all categorical values correctly for a higher number of data columns than Unique. The difference in their overall accuracies results from the inability of Unique to detect missing data. For example, the ``Chemox'' column of the CleanEHR dataset has three unique values: \texttt{0}, \texttt{1} and \texttt{NULL}. Here, the annotated categorical values are \texttt{0} and \texttt{1}, and \texttt{NULL} encodes missing data. While OpenML correctly labels \texttt{0} and \texttt{1} as the categorical values, Unique treats \texttt{NULL} as another categorical value. On the other hand, the Average Jaccard score shows that they provide similar coverage of categorical values per column. The main reason is that only a few data values are misclassified per column by Unique, which does not lead to large gaps in their performances. 

Next we test whether the competitor methods produce statistically different Jaccard indices per column than our method. Paired t-tests confirm that they are significantly different (see Appendix \ref{sec:add-exp-results-cat-val-inference} for the details). Additionally, we discuss the limitations of our method and how they can be mitigated in Appendix \ref{sec:additional-info-model}.

\pagebreak
 \bibliographystyle{splncs04}
 \bibliography{references}

\pagebreak
\appendix
\begin{center}
\textbf{\huge Supplemental Materials}
\end{center}

In this Supplemental Materials, we describe five data types (Appendix \ref{sec:defs-types}), and provide additional information about our model (Appendix \ref{sec:additional-info-model}) and our experimental setup (Appendix \ref{sec:additional-exp-setup}). Finally, we present additional discussion of the experimental results (Appendix \ref{sec:add-exp-results}).

\section{Definitions of Data Types}
\label{sec:defs-types}
The five main data types are described below:
\begin{itemize}
\item \emph{Categorical}. A categorical variable can take on
  one of a limited, and usually fixed, number of possible values \cite{yates2002practice}.
  A subtype is a Boolean variable, where there are only two possible values.
  The blood type of a person (\texttt{A}, \texttt{B}, \texttt{AB} or \texttt{O}) is an example of a
  non-Boolean categorical variable.

A problem for the identification of the categorical type from the data in a column is that the values may be \emph{encoded} in different ways. For the blood type example these are encoded as \emph{strings}, but the values might also be encoded as \emph{integers}, e.g., \texttt{1}, \texttt{2}, \texttt{3}, \texttt{4}. However, in both cases the limited number of possible values taken on by a categorical variable is the key to its identification. This problem may be eased if we have semantic knowledge about the variable (e.g., that it represents a blood type).

  In this classification we include ordinal variables (e.g.,
  poor, fair, good, excellent) as categorical variables, even though it
  may be useful to distinguish them further in subsequent analysis.
\item \emph{Date}: The  column contains date or date:time
  entries.  Considerable format variability is possible, e.g., from
  the  ISO-8601 format of \\
  \emph{YYYY-MM-DDThh:mm:ss} to e.g., 24 July 1969.
\item \emph{Float}: The column contains floating-point numbers. 
\item \emph{Integer}: The column contains integers. Note that we would regard the age of a person in
  years as an integer variable, but that the integer encoding of say
  blood type as \texttt{1}, \texttt{2}, \texttt{3} or \texttt{4} should be identified as a categorical variable.
\item \emph{String}: The string type is very general, and accommodates
data that do not fit into the four types above. It might contain
notes about a person, e.g., ``Works Mon-Wed mornings'', or a string
such as an IP address. However, a string encoding of
a categorical variable (e.g., blood types \texttt{A}, \texttt{B}, \texttt{AB} and \texttt{O}) should be
identified as being of categorical, not string type.
\end{itemize}
  
\section{Additional Information about Our Model}  
\label{sec:additional-info-model}
\begin{figure}[htb!]
    \centering
\includegraphics[width=\textwidth]{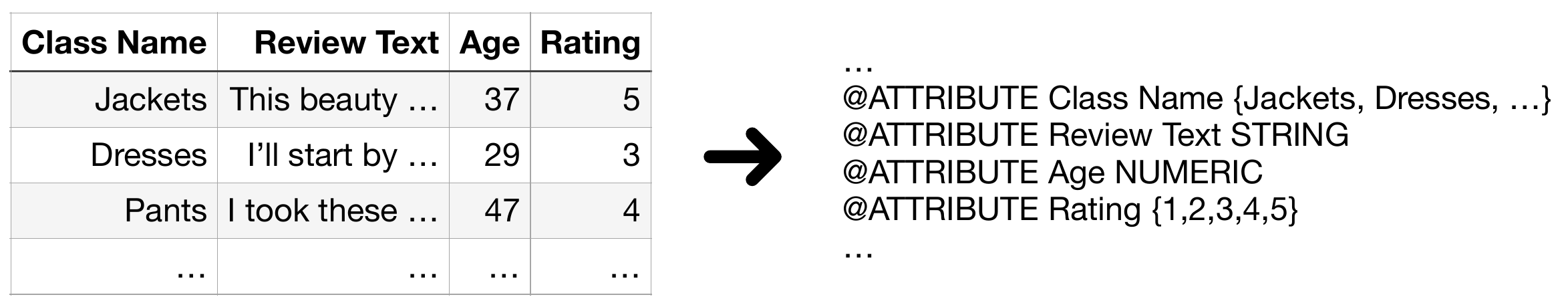}
    \captionof{figure}{A dataset and a section of the corresponding ARFF file.}
    \label{fig:ptype-cat-weka}
  \end{figure}

Following the notation used for ptype, we assume that a column of data $\textbf{x}=\{x_i\}_{i=1}^N$ has been read in as strings,
where each $x_i$ denotes the characters in the $i^{th}$ row. ptype is a generative model with a set of latent variables $t \in
\{1,2,...,K\}$ and $\textbf{z}=\{z_i\}_{i=1}^N$, where $t$ and $z_i$
respectively denote the data type of a column and its $i^{th}$
row. Here, $N$ is the number of rows in a data column and $K$ is the
number of possible data types for a column. The additional
missing and anomaly types are respectively denoted by $m$ and $a$. Note that $z_i$ can be of type $m$ or $a$ alongside
a regular data type, i.e. $z_i \in \{1,2,...,K,m,a\}$. This noisy
observation model allows a type inference procedure robustified for
missing and anomalous data values.

ptype-cat can enable the automatic construction of data dictionaries in the well-known Attribute Relation File Format (ARFF), which has been used notably to describe OpenML datasets including the UCI datasets (see Fig.\ \ref{fig:ptype-cat-weka} for an example).

\begin{figure}[htb!]
    \centering
\includegraphics[width=.5\textwidth]{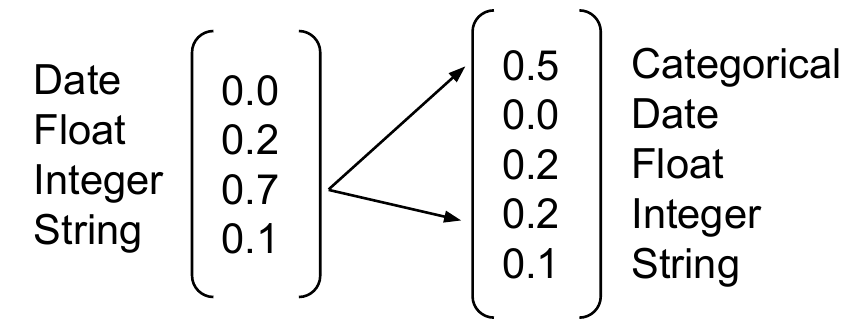}
    \captionof{figure}{A graphical representation of the re-distribution step where we split the probability mass for the integer type between the integer and categorical types.}
    \label{fig:ptype-cat-split}
  \end{figure}

Fig.~\ref{fig:ptype-cat-split} illustrates how the posterior probability for the integer type is divided between the integer and categorical types when the data type is initially inferred as integer.

\subsubsection*{Limitations of Our Model}
Below we discuss two limitations of our model and how they can be mitigated. Firstly, ptype-cat does not take into account the semantics of the categorical values, which can be important for data understanding. However, there is already a large number of works in the semantic web community that focus on extracting the semantics of the data, see e.g., \cite{chen2019colnet}. We believe that one could apply such methods once the categorical values are identified using our method. 

Secondly, our focus in this work has been the identification of the categorical values; however, a natural next step is to encode the categorical values for subsequent analysis. In this regard, ptype-cat cannot deal with ``dirty categories'' that occur due to string variability issues (e.g., MySQL and mysql would be labeled as two separate categorical values). We plan to resolve such issues through user interaction by letting the user merge dirty categories, although an alternative approach would be to directly encode them as vectors as in \emph{dirty-cat} \cite{cerda2018similarity}, where categorical values are encoded by taking into account string similarities. 
  
\section{Additional Information about Experimental Setup}
\label{sec:additional-exp-setup}
We describe the datasets, evaluation metrics and methods used in our experiments below:

\subsubsection*{Datasets}
For type inference, we have used 86 datasets obtained from various sources such as Kaggle\footnote{\url{https://www.kaggle.com/datasets}
[Accessed on 18/11/2020]}, OpenML\footnote{\url{https://https://www.openml.org/search?type=data}
[Accessed on 18/11/2020]} (randomly selected through the API) and
UCI\footnote{\url{https://archive.ics.uci.edu/ml/datasets.php}
[Accessed on 18/11/2020]}. We have annotated each dataset in terms of data types and categorical values by hand, based on the available meta-data and the unique values in each data column. The datasets are split using 5-fold nested cross-validation. Once the datasets are split, we collect their columns in the corresponding folds. Note that we split at the dataset level in order to avoid bias in test data. Our datasets contain a total number of 2989 columns (900 categorical, 49 date, 1462 float, 513 integer and 65 string). For the identification of categorical values task, we are interested in the 900 categorical data columns. Although 690 columns contain less than 6 unique categorical values, the number of categorical values in a column is between 1 and 80 (the data column with 80 categorical values, which is obtained from the CleanEHR dataset\footnote{The data is accessible via \url{https://github.com/ropensci/cleanEHR/tree/master/data} [Accessed on 04/12/2020].}, denotes the reason for a patient's admission following the ICNARC Coding Method\footnote{The details are available at \url{https://www.icnarc.org/Our-Audit/Audits/Cmp/Resources/Icm-Icnarc-Coding-Method}[Accessed on 04/12/2020].}).

\begin{algorithm}[htb!]
 \Data{X}
 \Model{M}
 \Parameters{P, K}
\For{$i=1$ \KwTo $K$}{
    Split X into $X_i^{training}$, $X_i^{test}$

    \For{$j=1$ \KwTo $K$}{
	    Split $X_i^{training}$ into $X_{ij}^{training}$, $X_{ij}^{test}$

     	\ForEach{$p \in P$}{
	    Train $M$ on $X_{ij}^{training}$

	    Calculate Error $E_{ijp}^{test}$
	    }
}
  Calculate Average Error $E_{ip}^{test}$

  Select $p^{*}$ where $E_{ip}^{test}$ is minimum

Train $M^{*}$ on $X_{i}^{training}$

Calculate Error $E_{i}^{test}$
    }
Calculate Average Error $E^{test}$
\vspace{.5cm}
 \caption{K-Fold Nested Cross-Validation}
\label{alg:nested-cv}
\end{algorithm}

\subsubsection*{Evaluation Metrics}
We use different sets of metrics for type inference and identification of categorical values. For type inference, we use the metrics which are used to evaluate ptype, namely overall accuracy and the Jaccard index. Note that Jaccard index allows us to measure the performance separately for each type following a \emph{one-vs-rest} approach. See \cite{ceritli2020ptype} for a detailed description of these metrics and how they are used for type inference. Additionally, we plot the Precision Recall (PR) curve for each method and report the Average Precision (AP) of each curve. These curves are obtained by micro-averaging over folds, meaning that the output probabilities of a method are concatenated across five outer folds of the nested cross-validation, and across the samples in each fold.

For identification of categorical values, we first evaluate the
methods using overall accuracy. The accuracy is 1 when the set of annotated categorical values is equal to the set of predicted categorical values and 0 otherwise. In order to take into account the partial matches between two sets, we calculate the Jaccard index per data column $J(A,B)$ defined as $|A \cap B| / |A \cup B|$, where $A$ and $B$ respectively denote the sets of annotated and predicted categorical values. Then we report their average over columns. 

\subsubsection*{Methods}
On type inference, we compare ptype-cat with Bot, OpenML and Weka. For Bot \cite{majoor2018}, we use the original implementation at \url{https://github.com/openml/ARFF-tools/blob/master/1030843_TheDataEncodingBot.ipynb}. Note that by default Bot considers only a subsample of a data column for computational efficiency. Here, we feed all the entries into the method in order to eliminate any bias. Additionally, we treat the pre-defined threshold for the number of unique values, which is 100 by default, as a hyperparameter. We use two different hyperparameters for the integer and string types to allow a wider search space since the lower thresholds are respectively 10 and 25. We estimate these parameters via nested cross-validation using a grid-search over the intervals of \{10, 20, \dots, 120\} and \{25, 35, \dots, 125\}. 

For ptype-cat, we use Logistic Regression with an L2 penalty with the regularization strength parameter selected in the interval of \{$10^{-4}$, \dots, $10^{4}$\}.

As we mention in Sec.\ \ref{sec:related_work}, the CSV to ARFF conversion methods used in OpenML and Weka are not directly applicable to our task. However, we adapt these methods by using ptype. We use ptype's prediction when a data column is labelled with the ARFF label numeric to classify the column either as float or integer. For OpenML's csv2arff method, we use the original implementation at \url{https://github.com/openml/ARFF-tools/blob/master/csv-to-arff.py} and treat the number of unique values as a hyperparameter. Although its default value is 10, we estimate this parameter via nested cross-validation using a grid-search over the interval of \{10, 20, \dots, 120\}. For Weka's method, we use the original implementation at \url{https://waikato.github.io/weka-wiki/formats_and_processing/converting_csv_to_arff/} which does not have any hyperparameters.

The methods adapted from OpenML and Weka can also be used for the identification of categorical values. Similarly, we adapt Bot to this task with a simple modification. Bot discards the data values that occur less than a threshold and does not treat them as categorical values. Here, we treat this threshold as a hyperparameter and estimate it via nested cross-validation using a grid-search over the interval of \{5, 10, 20, \dots, 80\}. In addition, we construct a baseline called Unique for inferring categorical values. Unique treats all unique values in a column as categorical values and is compared with ptype to demonstrate how much we can improve by eliminating missing and anomalous data.

\section{Additional Discussion about Experimental Results}
\label{sec:add-exp-results}

\paragraph{\bf{Type Inference:}}
\label{sec:add-exp-results-type-inference}
We obtain the hyperparameters of Bot as 10 and 25 respectively for the integer and string types and OpenML's unique values hyperparameter as 10 across test folds. 

\begin{figure}[h!]
\centering
\subfloat[Bot.]{\label{fig:mdleft}{\includegraphics[width=.25\textwidth]{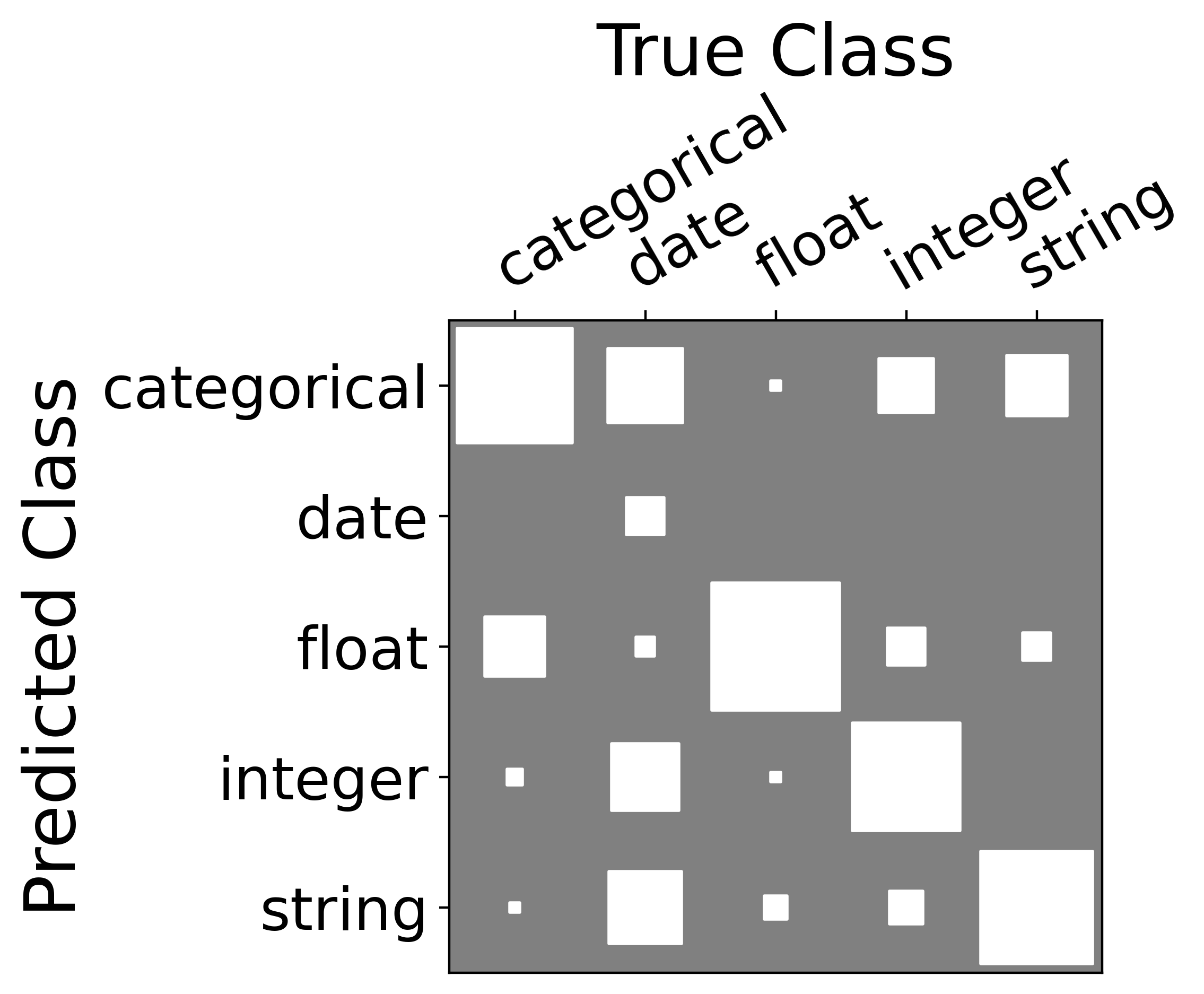}}}\hfill
\subfloat[OpenML.]{\label{fig:mdleft}{\includegraphics[width=.25\textwidth]{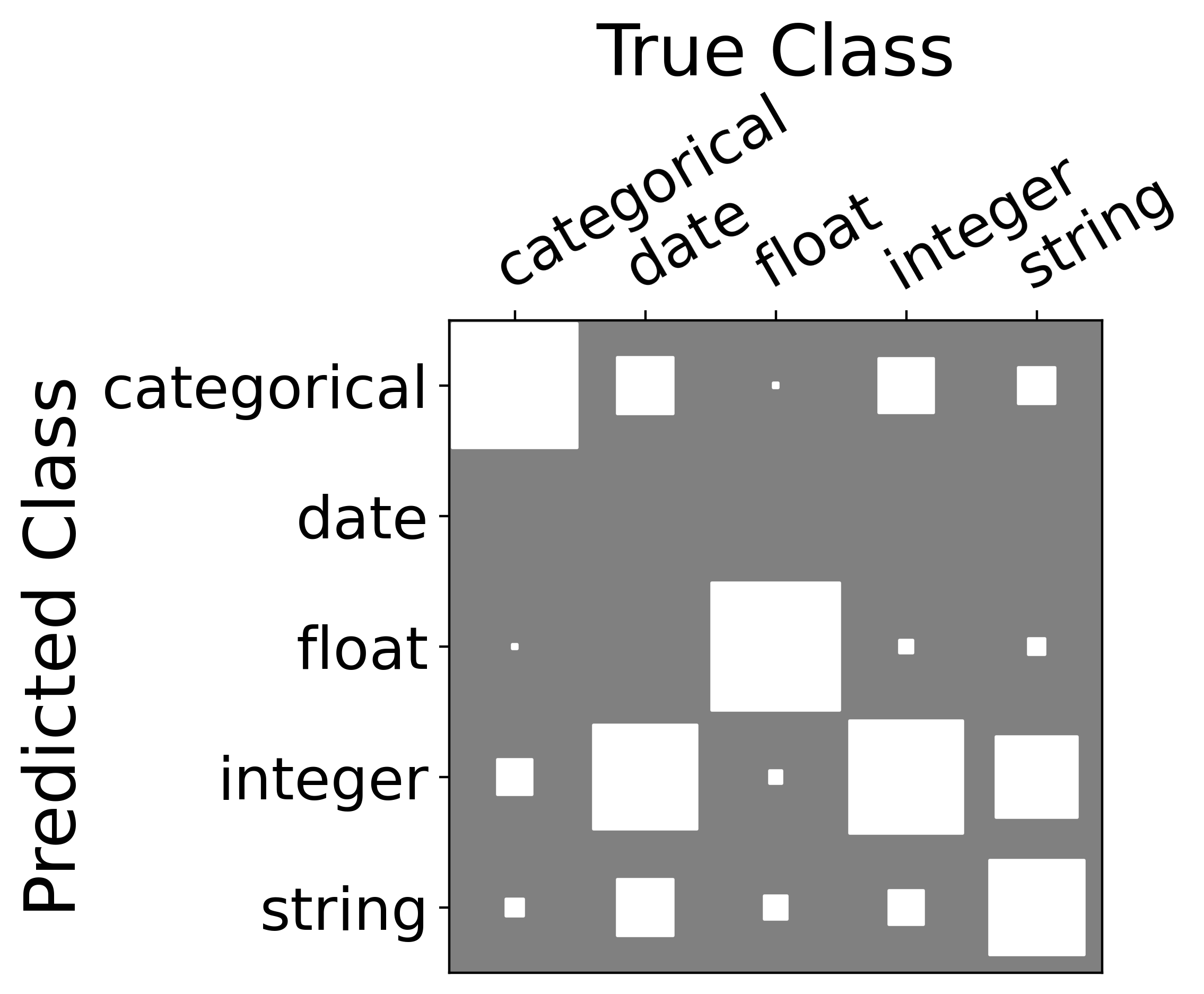}}}\hfill
\subfloat[Weka.]{\label{fig:mdleft}{\includegraphics[width=.25\textwidth]{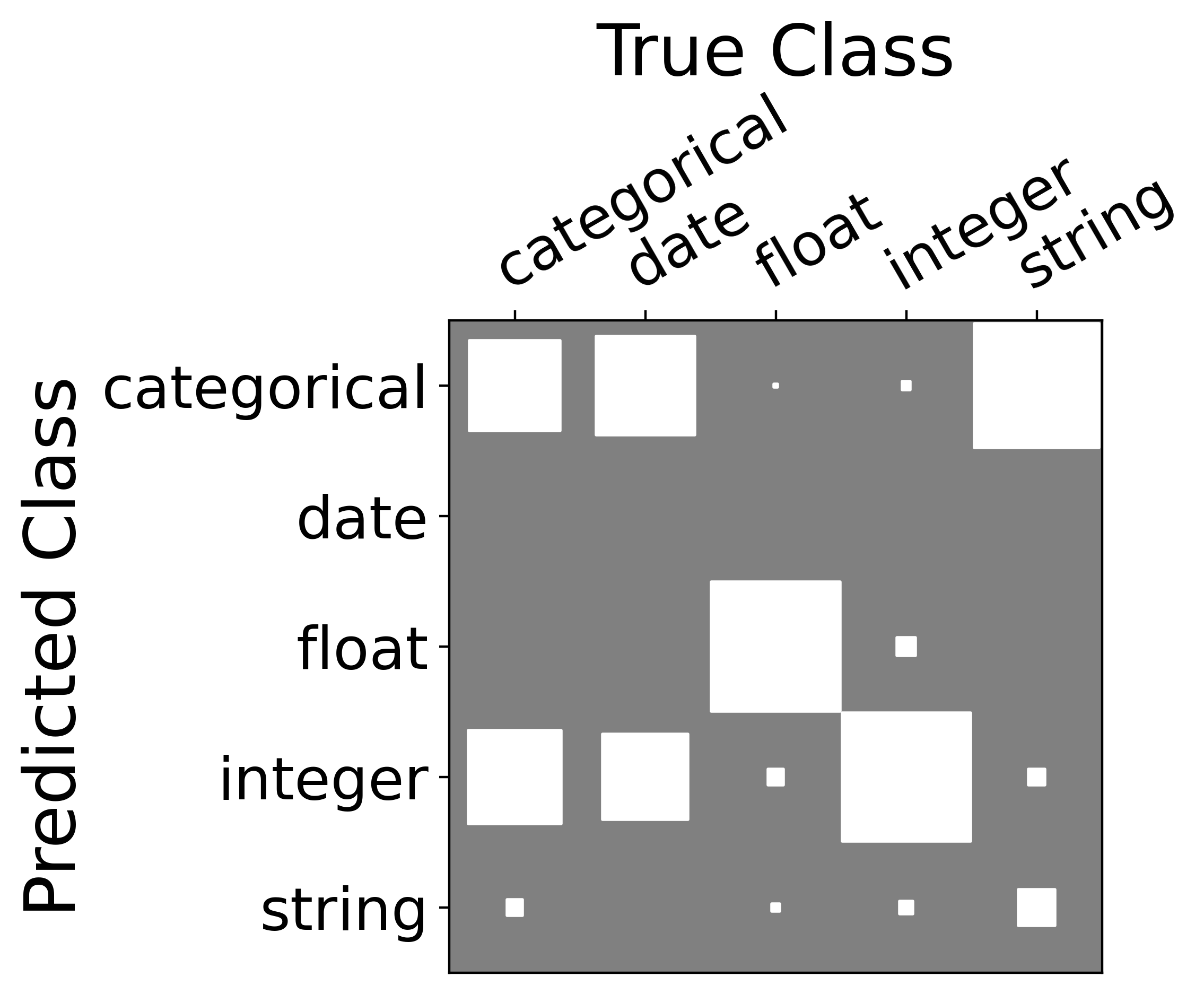}}}\hfill
\subfloat[ptype-cat.]{\label{fig:mdleft}{\includegraphics[width=.25\textwidth]{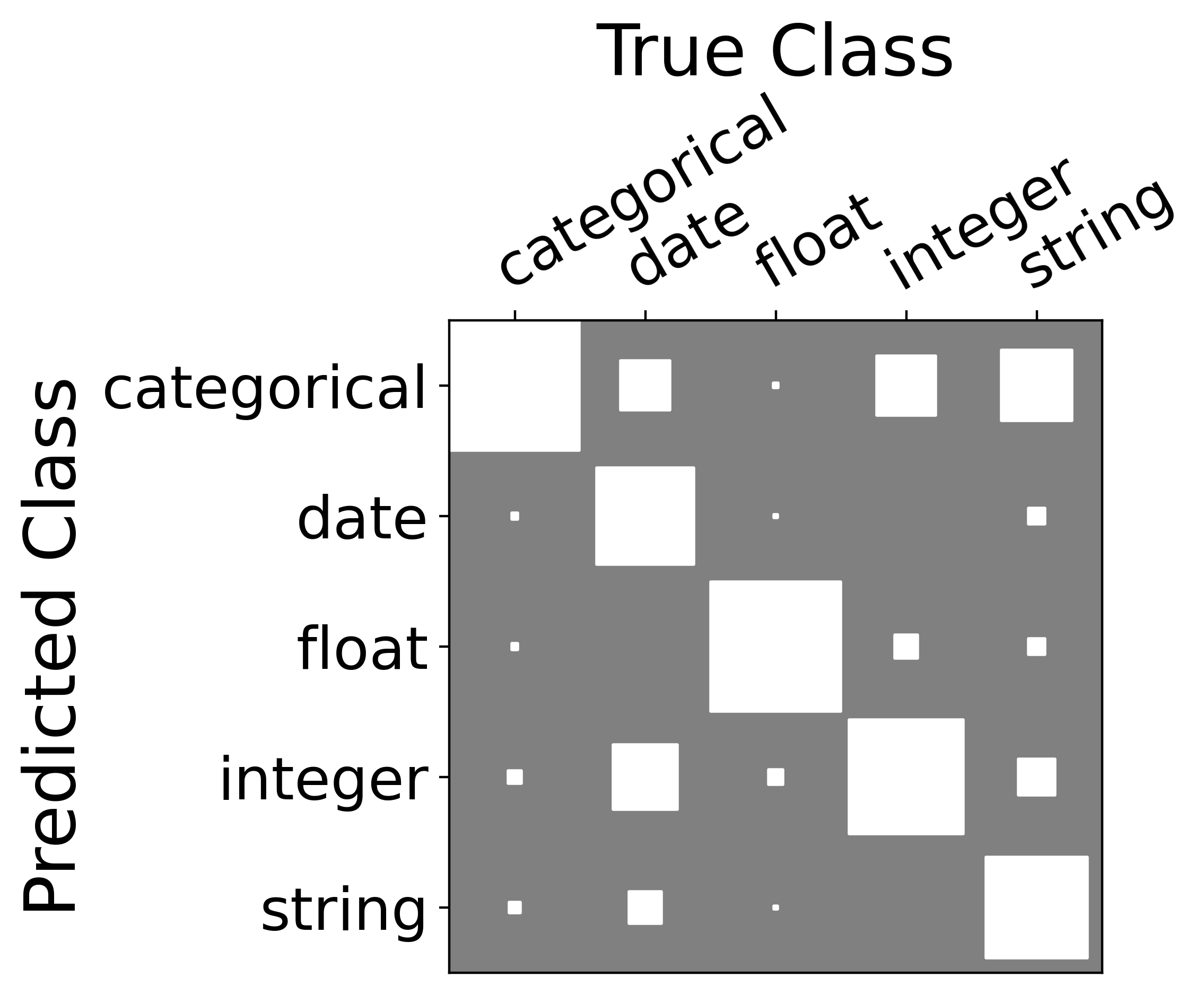}}}\hfill
\caption{Hinton plots of the normalized confusion matrices.}
\label{fig:hinton_diagrams}
\end{figure}

Fig.\ \ref{fig:hinton_diagrams} presents the normalized confusion matrices for the methods, normalized so that each column sums to 1. All methods lead to confusions by classifying columns as categorical rather than as integer or string. These failures are not surprising to some extent since categorical values are either encoded by integers or strings. However, ptype-cat has fewer such confusions than the others. 

ptype-cat performs better than the competitor methods for the date type, which can be explained by several reasons. The main reason for Bot is that it does not support date formats with time information. Instead, data columns in such formats are treated either as categorical or string, depending on the number of unique values. Additionally, Bot does not consider textual dates such as months and 4-digit formatted years, and results in misclassifications of categorical since the number of unique values is typically low. Weka considers the time information; however, it supports only the ISO-8601 format of ``yyyy-MM-ddTHH:mm:ss", unless the user specifies differently. Our method supports a more extensive set of formats including certain non-standard date formats thanks to the features obtained from ptype. OpenML completely discards the date type.

Unlike the other methods, Bot misclassifies a high number of categorical variables as floats. These failures occur when the Pandas library fails to parse a given dataset correctly. For example, the ``Active Sport'' column of the Young People Survey dataset---a categorical variable which ranks how active a young person is from \texttt{1} to \texttt{5}---is labelled with the float type by the Pandas library and consequently all integers are converted to their floating-point representations (e.g., \texttt{1.0} to \texttt{5.0}). Therefore, Bot is fed with these floating-point numbers rather than integers resulting in confusions. 

There are two main reasons to explain this difference between OpenML and ptype-cat. First, OpenML predicts the data type as integer when the number of unique values is higher than 10 (e.g., most columns of the Poker Hand dataset contain 13 categorical values encoded by integers). Secondly, OpenML does not properly handle missing data when the categorical values are encoded by strings. To correctly label such columns, it requires all the entries to be converted to string by using the Python's \emph{isistance} function which fails when the Pandas library detects missing data and encodes them as \texttt{np.nan}. For example, OpenML initially labels the ``HCMEST'' column of the
cleanEHR dataset (which contains values \texttt{Y}, \texttt{N} and \texttt{NULL}) as numeric rather than categorical. A forced choice between integer and float in 
ptype then results in equal posterior probabilities for these two types, 
as ptype must set its row latent variables to ``anomaly" to explain the
non-NULL data. 

To determine whether the column type predictions of ptype and OpenML
are significantly different, we apply McNemar's test
(see e.g.,  \cite{dietterich1998approximate}), which assumes that the two
methods should have the same error rate under the null hypothesis. 
We compute the test statistic
$(\abs{n_{01}-n_{10}})^2)/(n_{01}+n_{10})$, where $n_{01}$ and
$n_{10}$  denote 
the number of test columns misclassified by only OpenML, and by only
ptype respectively. In our case, $n_{01}$ and $n_{10}$ are respectively equal to 185 and 44,
which results in a statistic of 85.6. 
If the null hypothesis is correct, then the probability that this
statistic is greater than 3.84 is less
than 0.05 \cite{dietterich1998approximate}. Thus this result
provides evidence to reject the null hypothesis and confirms that
the methods are statistically significantly different from each other.

A common pattern in failure cases of all methods is that the assumption about the number of unique values does not always hold, i.e., there can be data columns of type integer or string with a low number of unique values. Consider the ``State'' column of the Geoplaces dataset which contains data values such as \texttt{Morelos}, \texttt{S.L.P.} and \texttt{San Luis Potosi}. The data column contains only 13 unique values out of 130 entries, which causes the methods to misclassify it as categorical (note that this data column is assumed to be of type string rather than categorical since the data is collected as free text). OpenML handles such cases slightly better than the others based on the heuristics used. However, its overall performance for the string type is poor as it classifies quite a high number of columns as integer rather than string due to the presence of missing data.

\paragraph{\bf{Identification of Categorical Values:}}
\label{sec:add-exp-results-cat-val-inference}
We obtain the hyperparameters of Bot as 5 across test folds. 

Bot obtains the worst performance, which is not surprising as it
relies on the number of occurrences of data values for detecting categorical values. Using the number of occurrences can become misleading for Bot. For example, it treats missing or anomalous data observed more than a pre-defined threshold in the data as categorical values. Similarly, it cannot identify categorical values that occur less than the same threshold.

Weka performs slightly better than Bot; however, its performance is still poor compared to the remaining methods. This is mainly because when it misclassifies a categorical variable, it generates an empty list for the corresponding categorical values, which causes both the accuracy and the Jaccard index to be zero.

In addition to the Average Jaccard score, we test whether methods produce statistically different Jaccard indices per column. We apply a paired t-test on the list of Jaccard indices obtained by Unique and ptype-cat (in the same order). Similarly, we obtain the p-values by OpenML-ptype-cat, Bot-ptype-cat and Weka-ptype-cat comparisions. We find that all the p-values are lower than $0.001$. These results reject the null hypothesis that the means are equal and confirm that they are significantly different.

\end{document}